%% file: Template.tex
\documentclass{article}
\usepackage{spconf,amsmath,graphicx,hyperref,amsfonts,booktabs,multirow}

\title{Decoder Design Matters for ECG Delineation}
\name{
\begin{tabular}{c}
Joseph Scharpf $^{*1}$, William Han$^{*1}$\thanks{* Equal contribution}, Chaojing Duan$^{2}$,\\
Michael A. Rosenberg$^{3}$, Emerson Liu$^{2}$, Ding Zhao$^{1}$
\end{tabular}
}
\address{$^{1}$Carnegie Mellon University, $^{2}$Allegheny Health Network, $^{3}$University of Colorado}
\begin{document}
%
\maketitle
\begin{abstract}
Electrocardiogram (ECG) delineation identifies the boundaries of P waves, QRS complexes, and T waves, providing structural annotations that can guide AI models in learning to interpret ECGs. However, training accurate delineation models requires manual annotations that are scarce and time-consuming to obtain. Recent work addresses this limitation through semi-supervised learning (SSL), but the design of the architecture, particularly the decoder, has received less attention. To this end, we propose \texttt{R-U-Net}, an ECG delineation model that pairs a ResNet-18 encoder with a U-Net decoder. On SemiSegECG, \texttt{R-U-Net} outperforms the strongest evaluated ResNet-18 + fully convolutional network (FCN) head baseline in each of the 16 in-domain settings by 3.3–13.0 mIoU and achieves 82.6 mIoU in the cross-domain setting, an improvement of 8.1 mIoU. Controlled ablations show that decoder design contributes more to performance gains than the evaluated SSL methods, motivating further exploration of architectures for ECG delineation. All code is open-source at \href{https://github.com/ELM-Research/ECG-Delineation}{\texttt{github.com/ELM-Research/ECG-Delineation}.}
\end{abstract}
\begin{keywords}
Electrocardiograms, ECG Delineation, Deep Learning, Semi-Supervised Learning
\end{keywords}
\input{sections/intro}
\input{sections/methods}
\input{sections/experimental}
\input{sections/results}
\input{sections/conclusion}

\vfill\pagebreak
\section{Compliance with Ethical Standards}
This study retrospectively analyzed publicly available, de-identified ECG data from LUDB~\cite{kalyakulina2020ludbnewopenaccessvalidation}, QTDB~\cite{qtdb}, ISP~\cite{avetisyan_2024_11472366}, Zhejiang~\cite{zheng2020ecgdatabase}, and PTB-XL~\cite{wagner_ptb-xl_2020}. No additional ethical approval was required for this secondary analysis.\\

\section{Acknowledgments}
This work was conducted in collaboration with the Mario Lemieux Center for Heart Rhythm Care at Allegheny General Hospital.
\bibliographystyle{IEEEbib}
\bibliography{strings,refs}

\end{document}

%% file: sections/intro.tex
\section{Introduction}
\label{sec:intro}
Applying artificial intelligence (AI) to interpret ECGs is a step towards scalability and automation.
While most recent works largely focus on tasks such as classification \cite{rajpurkar2017cardiologistlevelarrhythmiadetectionconvolutional} and ECG-conditioned language generation \cite{zhao2025ecgchatlargeecglanguagemodel}, ECG delineation has been a growing area of interest \cite{10.1145/3746252.3760790}.

ECG delineation is the task of partitioning the ECG in time into four categories: (1) background, (2) P wave, (3) QRS complex, and (4) T wave. Early automated ECG delineation methods relied on wavelet transforms and hand-designed rules \cite{martinez2004wavelet}. More recent approaches formulate delineation as dense sample-level labeling using architectures such as CNN-LSTM models \cite{peimankar2021dens} and 1D U-Net models \cite{jimenez2021unet}. Waveform annotations provide explicit ECG structure and granular supervision for downstream tasks such as ECG-conditioned language generation \cite{oh2026ecgreasoningbenchmarkbenchmarkevaluatingclinical}.
However, training accurate delineation models requires manual boundary annotations that are scarce and time-consuming to obtain.

To address this limitation, SemiSegECG \cite{10.1145/3746252.3760790} benchmarks semi-supervised learning (SSL) approaches for ECG delineation across six datasets.
Its comparison primarily focuses on SSL strategies, pairing ResNet \cite{he2015deepresiduallearningimage} and ViT \cite{dosovitskiy2021imageworth16x16words} encoders with a lightweight fully convolutional network (FCN) head.
Although U-Net architectures have previously been applied to ECG delineation, the contribution of decoder design within this benchmark remains unexplored.

In this study, we investigate decoder design for ECG delineation through \texttt{R-U-Net}, which pairs a ResNet-18 encoder with a U-Net decoder \cite{ronneberger2015unetconvolutionalnetworksbiomedical} in place of the FCN head.
\texttt{R-U-Net} outperforms the evaluated baselines in all 16 in-domain settings and the cross-domain setting of SemiSegECG \cite{10.1145/3746252.3760790}.
To isolate the contribution of decoder design, we conduct two ablation studies: (1) comparing decoder variants (Table~\ref{tab:decoder}) and (2) comparing SSL approaches (Table~\ref{tab:ssl}). In both studies, we find that the U-Net decoder contributes most to the performance gains. Altogether, these experiments highlight decoder design as a key factor in ECG delineation performance and motivate closer attention to architectural details.

%% file: sections/methods.tex
\begin{figure}[!t]
\centering
\includegraphics[width=0.9\columnwidth]{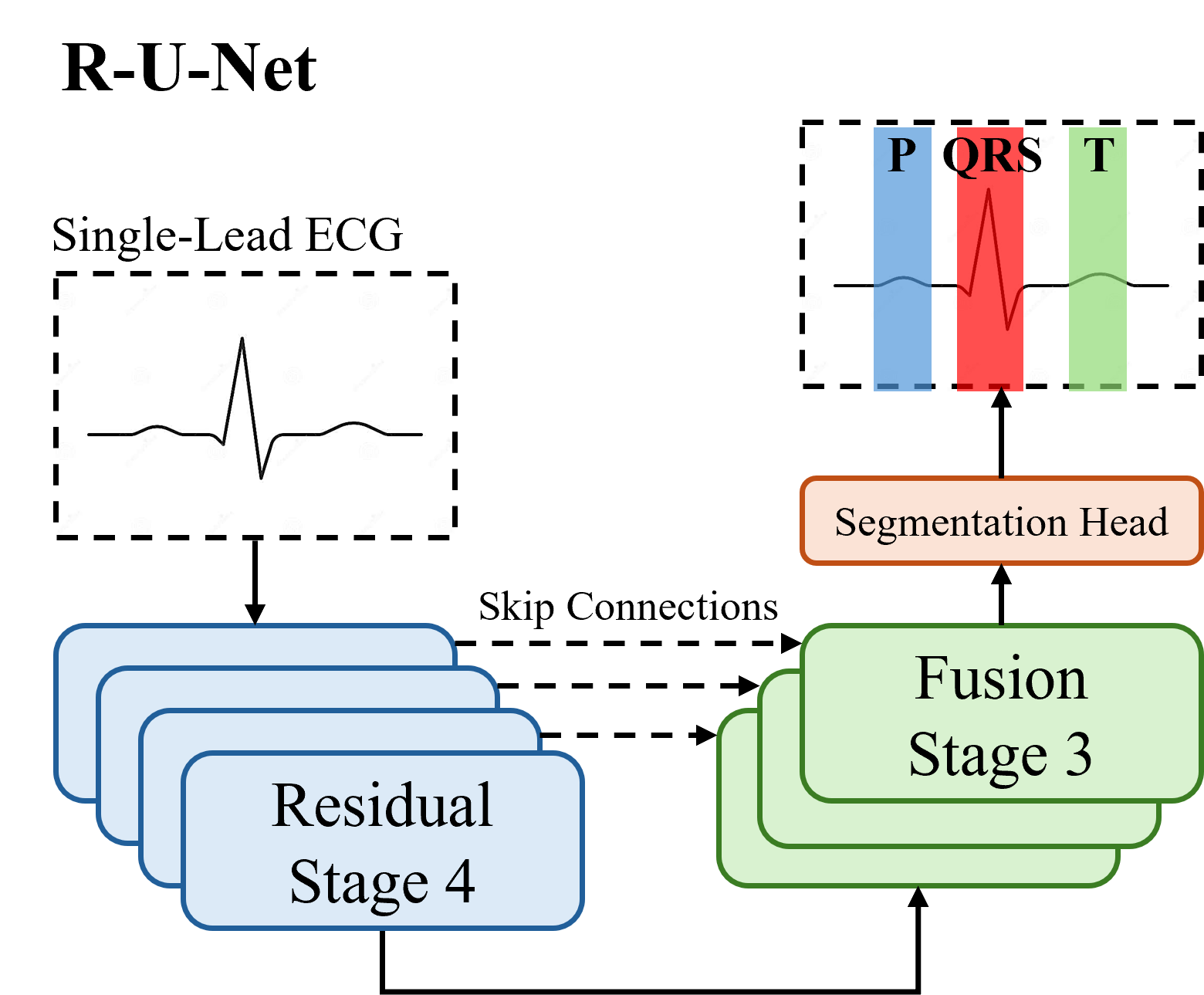}
\caption{A high-level architectural overview of \texttt{R-U-Net}.}
\label{fig:r-u-net}
\end{figure}
\begin{figure}[!t]
\centering
\includegraphics[width=1\columnwidth]{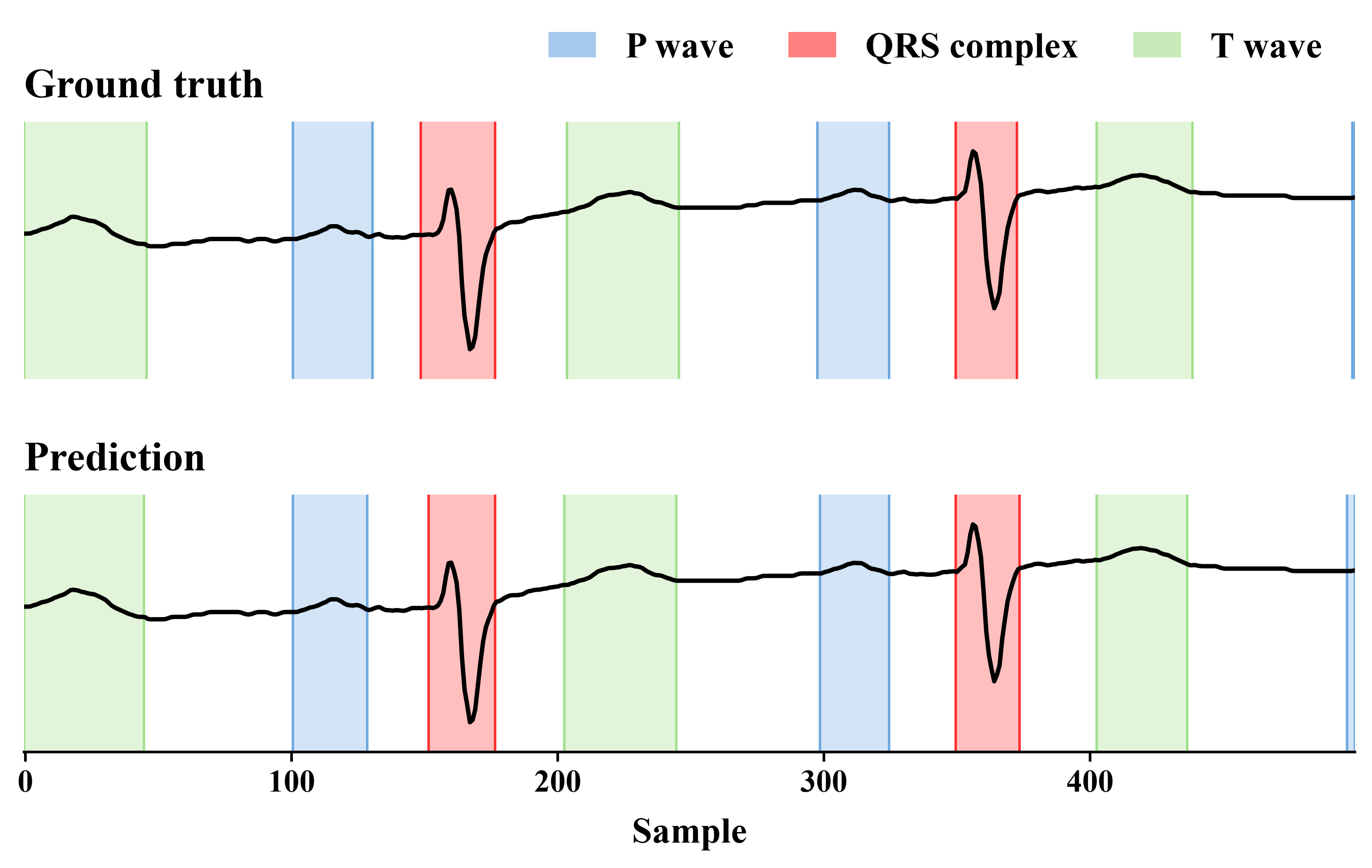}
\caption{Example of successful ECG delineation by \texttt{R-U-Net} compared with the ground truth, showing a 2-second segment from the middle of a 10-second ECG.}
\label{fig:qual}
\end{figure}
\section{Method}
\label{sec:method}

\subsection{Problem Formulation}
We follow SemiSegECG and formulate ECG delineation as sample-wise classification.
For a single-lead segment $x\in\mathbb{R}^{T}$, the target
$y\in\{0,1,2,3\}^{T}$ assigns each sample to background,
P wave, QRS complex, or T wave, respectively.
Given labeled segments $\mathcal{D}_{\mathrm{l}}$ and unlabeled
segments $\mathcal{D}_{\mathrm{u}}$, we learn a model $f_{\theta}$
whose output includes a class-wise softmax:
\begin{equation}
f_{\theta}(x)\in[0,1]^{4\times T},\qquad
\hat y_t=\arg\max_{c\in\{0,1,2,3\}}[f_{\theta}(x)]_{c,t}.
\label{eq:delineation}
\end{equation}
The first and last samples of each contiguous run of a
non-background class define its predicted onset and offset.

\subsection{\texttt{R-U-Net} Architecture Details}
\texttt{R-U-Net} combines a one-dimensional ResNet-18 encoder \cite{he2015deepresiduallearningimage} with a U-Net-style decoder \cite{ronneberger2015unetconvolutionalnetworksbiomedical}, both initialized from scratch. Figure~\ref{fig:r-u-net} provides an overview of the architecture.

The encoder contains four residual stages, each comprising two basic residual blocks. Given an augmented input $\widetilde{x}$, we retain the output of every stage:
\[
H_{\mathrm{enc}}
=
\left(
H_{\mathrm{enc}}^{(1)},\ldots,H_{\mathrm{enc}}^{(4)}
\right)
=
E_{\theta_{\mathrm{enc}}}(\widetilde{x}),
\]
where $H_{\mathrm{enc}}^{(s)}
\in \mathbb{R}^{d_s \times T_s}$.
The stage widths are $(64,128,256,512)$, with corresponding
temporal lengths $(625,313,157,79)$ for $T=2500$.

Starting from the deepest encoder representation, the decoder
progressively upsamples the features and concatenates them with
the corresponding encoder outputs through skip connections.
Each fusion stage applies two kernel-size-$3$ convolutions,
each followed by batch normalization and ReLU. The three stages
produce $256$, $128$, and $64$ channels, respectively.

The final decoder features are passed to a segmentation head.
Dropout with probability $0.1$ is followed by a pointwise
convolution that produces $C=4$ class logits. The head then
linearly interpolates these logits to the original signal
length and applies a class-wise softmax to obtain the predicted
class probabilities:
\[
f_{\theta}(\widetilde{x})
=
\operatorname{softmax}_{\mathrm{class}}
\left(
\mathcal{U}_{T}
\left(
D_{\theta_{\mathrm{dec}}}(H_{\mathrm{enc}})
\right)
\right),
\]
where $D_{\theta_{\mathrm{dec}}}$ comprises the decoder,
dropout, and pointwise classifier, and $\mathcal{U}_{T}$
denotes linear interpolation to length $T$.

\subsection{Boundary-Aware Mean Teacher}
We use Mean Teacher~\cite{meanteacher} with a student
$f_{\theta}$ and a teacher $f_{\bar\theta}$. The teacher is initialized
from the student and updated after each optimization step as
$\bar\theta\leftarrow0.99\bar\theta+0.01\theta$.
It operates in evaluation mode without gradient updates.
For each unlabeled segment, the teacher receives a weak view
$x^{\mathrm{w}}$ obtained by random temporal resizing and
padding or cropping. The student receives a strong view
$x^{\mathrm{s}}$ with additional stochastic amplitude
perturbations and powerline, white, or sinusoidal noise.
The two views remain temporally aligned. Labeled segments
and their masks undergo the same weak transformation.

For a labeled minibatch $\mathcal{B}_{\mathrm{l}}$, we use
sample-wise cross-entropy:
\begin{equation}
\mathcal{L}_{\mathrm{sup}}
=-\frac{1}{|\mathcal{B}_{\mathrm{l}}|T}
\sum_{(x,y)\in\mathcal{B}_{\mathrm{l}}}\sum_{t=1}^{T}
\log[f_{\theta}(x)]_{y_t,t},
\label{eq:supervised}
\end{equation}
where $(x,y)$ denotes an augmented segment and its aligned mask.
For each unlabeled segment, write
$q=f_{\bar\theta}(x^{\mathrm{w}})$ and
$p=f_{\theta}(x^{\mathrm{s}})$.
Both consistency terms use the soft-target cross-entropy
$\ell_t=-\sum_{c=0}^{3}q_{c,t}\log p_{c,t}$.

To emphasize potential waveform boundaries, we measure
changes between adjacent teacher probability vectors:
\begin{equation}
\delta_t=\frac{1}{2}\sum_{c=0}^{3}|q_{c,t}-q_{c,t-1}|,
\qquad t=2,\ldots,T,
\label{eq:probability_change}
\end{equation}
with $\delta_1=0$. We spread these changes over a $\pm4$-sample
neighborhood and normalize over the full segment:
\begin{equation}
e_t=\max_{\substack{1\le u\le T\\|u-t|\le4}}\delta_u,
\qquad
b_t=\frac{e_t}{\max(10^{-6},\max_u e_u)}.
\label{eq:boundary_map}
\end{equation}
Let $a_t=\max_c q_{c,t}$ denote teacher confidence.
Set $g=1$ if its mean over the segment is at least $0.50$,
and $g=0$ otherwise.
The region and boundary weights are
\begin{equation}
\begin{aligned}
w_t^{\mathrm{r}}&=g(1-b_t)\mathbf{1}[a_t\ge0.80],\\
w_t^{\mathrm{b}}&=g b_t(1+a_t)/2.
\end{aligned}
\label{eq:consistency_weights}
\end{equation}
The region term favors confident positions away from
probability changes, including background. The boundary term
emphasizes these changes without the hard sample-wise
confidence threshold.

For an unlabeled minibatch $\mathcal{B}_{\mathrm{u}}$, we
normalize the two terms independently:
\begin{equation}
\mathcal{L}_{k}=
\frac{\sum_{x\in\mathcal{B}_{\mathrm{u}}}\sum_{t=1}^{T}
w_t^{k}\ell_t}
{\max\!\left(1,\sum_{x\in\mathcal{B}_{\mathrm{u}}}
\sum_{t=1}^{T}w_t^{k}\right)},
\quad k\in\{\mathrm{r},\mathrm{b}\},
\label{eq:weighted_consistency}
\end{equation}
where the dependence of $w_t^{k}$ and $\ell_t$ on $x$ is
implicit. The student minimizes
\begin{equation}
\mathcal{L}=\frac{1}{2}\left(
\mathcal{L}_{\mathrm{sup}}+\mathcal{L}_{\mathrm{r}}
+\mathcal{L}_{\mathrm{b}}\right).
\label{eq:training_objective}
\end{equation}

%% file: sections/experimental.tex
\section{Experimental Settings}

\subsection{Datasets}
We follow the public in-domain and merged cross-domain protocols of SemiSegECG~\cite{10.1145/3746252.3760790}, using its supplied training, validation, and test splits. LUDB~\cite{kalyakulina2020ludbnewopenaccessvalidation}, QTDB~\cite{qtdb}, ISP~\cite{avetisyan_2024_11472366}, and Zhejiang~\cite{zheng2020ecgdatabase} provide delineation labels. Each lead is treated as an independent input. Under the in-domain protocol (Table~\ref{tab:main}), labeled and unlabeled data come from the same dataset. Random subsets comprising 1/16, 1/8, 1/4, or 1/2 of the training set serve as labeled data, while the entire training set serves as unlabeled data. Under the merged cross-domain protocol (Figure~\ref{fig:cross}), the four labeled datasets are combined while preserving their original splits, and PTB-XL~\cite{wagner_ptb-xl_2020} serves as an external unlabeled dataset. We evaluate on the merged in-domain test set to assess performance when labeled and unlabeled training data come from different sources. Following SemiSegECG preprocessing, waveforms are resampled to 250 Hz, producing $T=2500$ samples, and processed with high-pass and low-pass filters at 0.67 and 40\,Hz. Z-score normalization is applied to all model inputs.

\subsection{Training and Evaluation}
We train for 100 epochs using AdamW with a learning rate of
$10^{-3}$ and weight decay of 0.05. The learning rate increases
linearly during the first 10 epochs and subsequently follows
a cosine schedule toward $10^{-4}$. Each step uses 16 labeled
and 16 unlabeled examples. We select the student checkpoint
with the highest validation mean intersection-over-union
(mIoU) and evaluate it on the test set. mIoU includes
all four classes, including background. All mIoU results with standard deviations are from our experiments and are reported as the mean \textpm{} standard deviation across three random seeds.

%% file: sections/results.tex
\input{tables/table1}
\section{Results}
\label{sec:results}

\subsection{In-Domain Evaluation}
Table~\ref{tab:main} presents the 16 in-domain evaluations on SemiSegECG. We compare \texttt{R-U-Net} trained using Boundary-aware MT with six baseline training methods using the ResNet-18 + FCN architecture, as reported in SemiSegECG~\cite{10.1145/3746252.3760790}. \texttt{R-U-Net} achieves the highest mIoU across all four datasets and four labeled-data proportions, outperforming the strongest baseline in each setting by 3.3 to 13.0 mIoU. At the lowest labeled-data proportion (1/16), the improvements are 13.0, 8.9, 12.3, and 4.5 mIoU on LUDB, QTDB, ISP, and Zhejiang, respectively. These results demonstrate consistent improvements across datasets and levels of labeled-data availability, including settings with limited annotations.

\begin{figure}[!t]
\centering
\includegraphics[width=\columnwidth]{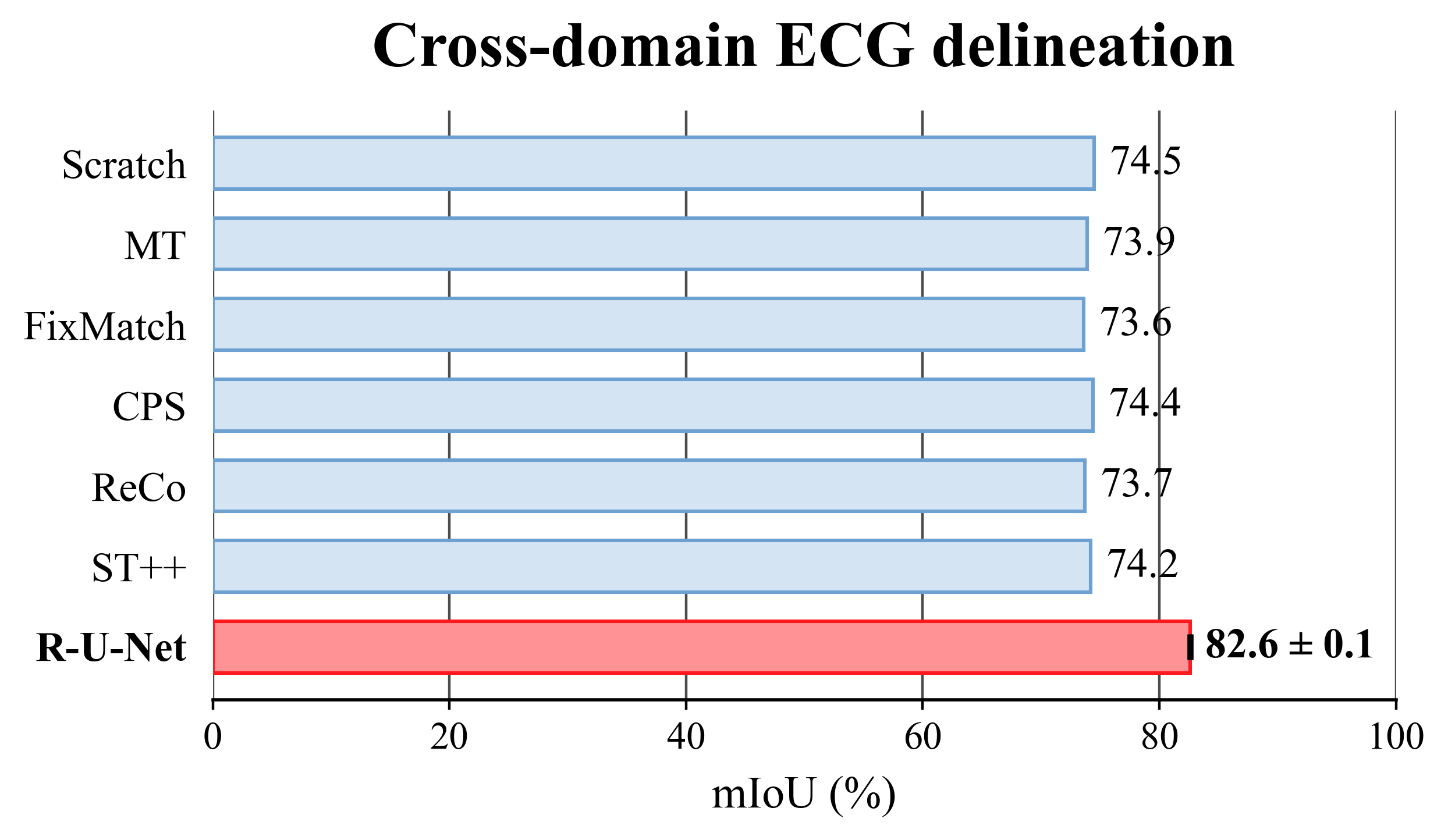}
\caption{Cross-domain test mIoU on the merged in-domain test set. Baselines use an FCN decoder~\cite{10.1145/3746252.3760790}.}
\label{fig:cross}
\end{figure}

\subsection{Cross-Domain Evaluation}
Figure~\ref{fig:cross} compares \texttt{R-U-Net} trained using Boundary-aware MT with six baseline training methods using the ResNet-18 + FCN architecture in the cross-domain setting. \texttt{R-U-Net} achieves the highest mIoU of 82.6, exceeding the strongest baseline, Scratch (74.5 mIoU), by 8.1 mIoU. These results show that \texttt{R-U-Net} maintains its performance advantage when labeled and unlabeled training data come from different sources, supporting its effectiveness beyond the in-domain training setting.

\subsection{Comparing Decoder Variants}
\input{tables/table2}
We evaluate two intermediate decoder variants: (1) \textbf{Wide FCN}, which adds an additional convolutional layer and increases the channel widths to match the parameter count of our U-Net decoder ($\sim$1.03M), and (2) \textbf{U-Net (w/o skip connections)}, which is the same U-Net decoder used in \texttt{R-U-Net} without skip connections. To isolate the effect of the decoder, we train Wide FCN, U-Net w/o skip connections, and U-Net with the scratch training method. Wide FCN achieves comparable performance to the original FCN (67.2 versus 67.3 mIoU). The U-Net decoder without skip connections achieves 80.0 mIoU, while adding skip connections improves mIoU by 1.8 points to 81.8. These results suggest that the gains primarily arise from the U-Net decoder, with skip connections providing a minor increase.

\subsection{Comparing SSL Approaches}
\input{tables/table3}
Table~\ref{tab:ssl} compares different training methods for \texttt{R-U-Net} with the ResNet-18 + FCN baselines on LUDB 1/16. Under scratch training, replacing the FCN decoder with the U-Net decoder increases mIoU from 67.3 to 81.8, a gain of 14.5 mIoU. Scratch \texttt{R-U-Net} also exceeds the strongest FCN baseline, ReCo (71.5 mIoU), by 10.3 mIoU. SSL provides smaller additional gains: FixMatch, standard MT, and boundary-aware MT achieve 83.5, 84.0, and 84.5 mIoU, respectively. Boundary-aware MT improves over scratch training by 2.7 mIoU and standard MT by only 0.5 mIoU. These results indicate that decoder design accounts for most of the improvement in this setting, with SSL providing modest additional gains.

%% file: tables/table1.tex

\begin{table}[!t]
\centering
\small
\caption{In-domain test mIoU (\%) under varying labeled-data ratios. Baseline results are from SemiSegECG \cite{10.1145/3746252.3760790} and use a ResNet-18 + FCN. \texttt{R-U-Net} results are reported as mean\textpm{}standard deviation over three seeds.}
\label{tab:main}
\resizebox{\columnwidth}{!}{%
\begin{tabular}{lcccc}
\toprule
\textbf{Method}
  & \multicolumn{4}{c}{\textbf{ResNet-18}} \\
\cmidrule(lr){2-5}
  & 1/16 & 1/8 & 1/4 & 1/2 \\
\midrule

\multicolumn{5}{c}{\textbf{LUDB}} \\
\addlinespace[2pt]
Scratch  & 67.3 & 71.3 & 72.9 & 74.1 \\
MT       & 70.8 & 72.3 & 73.6 & 74.3 \\
FixMatch & 70.9 & 72.2 & 72.9 & 74.1 \\
CPS      & 68.6 & 71.6 & 73.1 & 74.3 \\
ReCo     & 71.5 & 72.5 & 73.1 & 73.9 \\
ST++     & 69.2 & 71.6 & 73.7 & 74.5 \\
\addlinespace[2pt]
\texttt{R-U-Net}
  & \textbf{84.5\textpm{}0.1} & \textbf{85.5\textpm{}0.1}
  & \textbf{85.5\textpm{}0.3} & \textbf{85.9\textpm{}0.2} \\
\midrule

\multicolumn{5}{c}{\textbf{QTDB}} \\
\addlinespace[2pt]
Scratch  & 47.5 & 56.2 & 60.5 & 64.9 \\
MT       & 47.8 & 44.8 & 63.0 & 66.7 \\
FixMatch & 46.7 & 53.3 & 58.2 & 66.3 \\
CPS      & 53.2 & 57.1 & 64.7 & 68.0 \\
ReCo     & 53.4 & 53.4 & 58.7 & 64.5 \\
ST++     & 52.9 & 57.8 & 62.5 & 68.0 \\
\addlinespace[2pt]
\texttt{R-U-Net}
  & \textbf{62.3\textpm{}3.0} & \textbf{67.7\textpm{}0.5}
  & \textbf{75.1\textpm{}0.1} & \textbf{76.8\textpm{}0.6} \\
\midrule

\multicolumn{5}{c}{\textbf{ISP}} \\
\addlinespace[2pt]
Scratch  & 62.3 & 64.6 & 68.1 & 69.3 \\
MT       & 64.2 & 65.9 & 68.0 & 69.1 \\
FixMatch & 63.1 & 64.4 & 67.9 & 68.9 \\
CPS      & 64.4 & 66.1 & 68.5 & 69.4 \\
ReCo     & 62.3 & 64.7 & 67.4 & 68.1 \\
ST++     & 63.7 & 65.7 & 68.5 & 69.5 \\
\addlinespace[2pt]
\texttt{R-U-Net}
  & \textbf{76.7\textpm{}0.2} & \textbf{77.9\textpm{}0.2}
  & \textbf{79.3\textpm{}0.2} & \textbf{79.9\textpm{}0.1} \\
\midrule

\multicolumn{5}{c}{\textbf{Zhejiang}} \\
\addlinespace[2pt]
Scratch  & 76.7 & 78.9 & 80.7 & 82.3 \\
MT       & 79.2 & 80.1 & 81.5 & 82.9 \\
FixMatch & 79.0 & 80.4 & 81.2 & 82.7 \\
CPS      & 77.6 & 79.5 & 81.3 & 82.7 \\
ReCo     & 77.6 & 78.9 & 79.7 & 80.6 \\
ST++     & 78.8 & 79.8 & 81.9 & 82.9 \\
\addlinespace[2pt]
\texttt{R-U-Net}
  & \textbf{83.7\textpm{}0.1} & \textbf{84.8\textpm{}0.1}
  & \textbf{85.4\textpm{}0.0} & \textbf{86.2\textpm{}0.1} \\
\bottomrule
\end{tabular}}
\end{table}

%% file: tables/table2.tex
\begin{table}[!t]
\centering
\small
\caption{Ablation study of decoder variants on the LUDB dataset \cite{kalyakulina2020ludbnewopenaccessvalidation} using a labeled-data proportion of 1/16.}
\label{tab:decoder}
\resizebox{0.7\columnwidth}{!}{%
\begin{tabular}{lc}
\toprule
Decoder & mIoU \\
\midrule
FCN & 67.3 \\
Wide FCN & 67.2\textpm{}0.3 \\
U-Net (w/o skip connections) & 80.0\textpm{}0.1 \\
U-Net & \textbf{81.8\textpm{}0.2} \\
\bottomrule
\end{tabular}}
\end{table}

%% file: tables/table3.tex
\begin{table}[!t]
\centering
\small
\caption{Ablation study on applying different semi-supervised training methods to \texttt{R-U-Net}. We conduct experiments on LUDB 1/16.}
\label{tab:ssl}
\resizebox{\columnwidth}{!}{%
\begin{tabular}{llc}
\toprule
Architecture & Method & mIoU \\
\midrule
\multirow{6}{*}{ResNet-18 + FCN}
 & Scratch & 67.3 \\
 & MT & 70.8 \\
 & FixMatch & 70.9 \\
 & CPS & 68.6 \\
 & ReCo & 71.5 \\
 & ST++ & 69.2 \\
\midrule
\multirow{4}{*}{\texttt{R-U-Net}}
 & Scratch & 81.8\textpm{}0.2 \\
 & MT & 84.0\textpm{}0.1 \\
 & FixMatch & 83.5\textpm{}0.2 \\
 & Boundary-aware MT (Ours) & \textbf{84.5\textpm{}0.1} \\
\bottomrule
\end{tabular}}
\end{table}

%% file: sections/conclusion.tex
\section{Conclusion}
\label{sec:conclusion}
In this paper, we introduce \texttt{R-U-Net}, which pairs a ResNet-18 encoder with a U-Net decoder for ECG delineation. Across the SemiSegECG benchmark, \texttt{R-U-Net} outperforms the evaluated ResNet-18 + FCN baselines in all 16 in-domain settings and the cross-domain setting, demonstrating consistent improvements across datasets and levels of labeled-data availability. Controlled ablations indicate that decoder design accounts for most of these gains. Increasing the FCN decoder's parameter count provides no improvement, whereas the U-Net decoder substantially improves performance even without skip connections. Under scratch training, \texttt{R-U-Net} also exceeds the strongest evaluated FCN baseline by 10.3 mIoU points. Together, these findings highlight the importance of establishing strong architectural baselines when assessing SSL for ECG delineation, particularly under limited supervision. Future work could examine whether these decoder-level findings extend to other encoders and datasets, and develop SSL methods that further improve delineation when annotations are scarce.

%% file: refs.bib
@article{peimankar2021dens,
  author  = {Peimankar, Abdolrahman and Puthusserypady, Sadasivan},
  title   = {{DENS-ECG}: A Deep Learning Approach for {ECG} Signal Delineation},
  journal = {Expert Systems with Applications},
  year    = {2021},
  volume  = {165},
  note    = {Art. no. 113911},
  doi     = {10.1016/j.eswa.2020.113911}
}

@article{jimenez2021unet,
  author  = {Jimenez-Perez, Guillermo and Alcaine, Alejandro and Camara, Oscar},
  title   = {Delineation of the Electrocardiogram with a Mixed-Quality-Annotations Dataset Using Convolutional Neural Networks},
  journal = {Scientific Reports},
  year    = {2021},
  volume  = {11},
  note    = {Art. no. 863},
  doi     = {10.1038/s41598-020-79512-7}
}

@article{martinez2004wavelet,
  author  = {Mart{\'\i}nez, Juan Pablo and Almeida, Rute and Olmos, Salvador and Rocha, Ana Paula and Laguna, Pablo},
  title   = {A Wavelet-Based {ECG} Delineator: Evaluation on Standard Databases},
  journal = {{IEEE} Transactions on Biomedical Engineering},
  year    = {2004},
  volume  = {51},
  number  = {4},
  pages   = {570--581},
  doi     = {10.1109/TBME.2003.821031}
}

@misc{ronneberger2015unetconvolutionalnetworksbiomedical,
      title={U-Net: Convolutional Networks for Biomedical Image Segmentation}, 
      author={Olaf Ronneberger and Philipp Fischer and Thomas Brox},
      year={2015},
      eprint={1505.04597},
      archivePrefix={arXiv},
      primaryClass={cs.CV},
      url={https://arxiv.org/abs/1505.04597}, 
}

@misc{dosovitskiy2021imageworth16x16words,
      title={An Image is Worth 16x16 Words: Transformers for Image Recognition at Scale}, 
      author={Alexey Dosovitskiy and Lucas Beyer and Alexander Kolesnikov and Dirk Weissenborn and Xiaohua Zhai and Thomas Unterthiner and Mostafa Dehghani and Matthias Minderer and Georg Heigold and Sylvain Gelly and Jakob Uszkoreit and Neil Houlsby},
      year={2021},
      eprint={2010.11929},
      archivePrefix={arXiv},
      primaryClass={cs.CV},
      url={https://arxiv.org/abs/2010.11929}, 
}

@misc{he2015deepresiduallearningimage,
      title={Deep Residual Learning for Image Recognition}, 
      author={Kaiming He and Xiangyu Zhang and Shaoqing Ren and Jian Sun},
      year={2015},
      eprint={1512.03385},
      archivePrefix={arXiv},
      primaryClass={cs.CV},
      url={https://arxiv.org/abs/1512.03385}, 
}

@inproceedings{meanteacher,
  author    = {Tarvainen, Antti and Valpola, Harri},
  title     = {Mean teachers are better role models: Weight-averaged consistency targets improve semi-supervised deep learning results},
  booktitle = {Advances in Neural Information Processing Systems},
  year      = {2017},
  volume    = {30}
}

@misc{avetisyan_2024_11472366,
  author       = {Avetisyan, Aram and
                  Khachaturov, Nikolas and
                  Asatryan, Ariana and
                  Tigranyan, Shahane and
                  Markin, Yury},
  title        = {ISP ECG delineation dataset},
  month        = jun,
  year         = 2024,
  publisher    = {Zenodo},
  doi          = {10.5281/zenodo.11472366},
  url          = {https://doi.org/10.5281/zenodo.11472366},
}

@article{wagner_ptb-xl_2020,
        title = {{PTB}-{XL}, a large publicly available electrocardiography dataset},
        volume = {7},
        copyright = {2020 The Author(s)},
        issn = {2052-4463},
        url = {https://www.nature.com/articles/s41597-020-0495-6},
        doi = {10.1038/s41597-020-0495-6},
        language = {en},
        number = {1},
        urldate = {2020-10-28},
        journal = {Scientific Data},
        author = {Wagner, Patrick and Strodthoff, Nils and Bousseljot, Ralf-Dieter and Kreiseler, Dieter and Lunze, Fatima I. and Samek, Wojciech and Schaeffter, Tobias},
        month = may,
        year = {2020},
        note = {Number: 1
Publisher: Nature Publishing Group},
        pages = {154}
}

@article{zheng2020ecgdatabase,
  title   = {A 12-Lead {ECG} database to identify origins of idiopathic ventricular arrhythmia containing 334 patients},
  author  = {Zheng, Jianwei and Fu, Guohua and Anderson, Kyle and Chu, Huimin and Rakovski, Cyril},
  journal = {Scientific Data},
  volume  = {7},
  number  = {1},
  pages   = {98},
  year    = {2020},
  doi     = {10.1038/s41597-020-0440-8}
}

@article{qtdb,
author = {Laguna, Pablo and Mark, R.G. and Goldberg, A. and Moody, G.B.},
year = {1997},
month = {10},
pages = {673 - 676},
title = {Database for evaluation of algorithms for measurement of QT and other waveform intervals in the ECG},
volume = {1997},
journal = {Computers in Cardiology},
doi = {10.1109/CIC.1997.648140}
}

@misc{kalyakulina2020ludbnewopenaccessvalidation,
      title={LUDB: a new open-access validation tool for electrocardiogram delineation algorithms}, 
      author={Alena I. Kalyakulina and Igor I. Yusipov and Victor A. Moskalenko and Alexander V. Nikolskiy and Konstantin A. Kosonogov and Grigory V. Osipov and Nikolai Yu. Zolotykh and Mikhail V. Ivanchenko},
      year={2020},
      eprint={1809.03393},
      archivePrefix={arXiv},
      primaryClass={q-bio.QM},
      url={https://arxiv.org/abs/1809.03393}, 
}

@misc{zhao2025ecgchatlargeecglanguagemodel,
      title={ECG-Chat: A Large ECG-Language Model for Cardiac Disease Diagnosis}, 
      author={Yubao Zhao and Jiaju Kang and Tian Zhang and Puyu Han and Tong Chen},
      year={2025},
      eprint={2408.08849},
      archivePrefix={arXiv},
      primaryClass={eess.SP},
      url={https://arxiv.org/abs/2408.08849}, 
}

@inproceedings{10.1145/3746252.3760790,
author = {Park, Minje and Lim, Jeonghwa and Yu, Taehyung and Joo, Sunghoon},
title = {SemiSegECG: A Multi-Dataset Benchmark for Semi-Supervised Semantic Segmentation in ECG Delineation},
year = {2025},
isbn = {9798400720406},
publisher = {Association for Computing Machinery},
address = {New York, NY, USA},
url = {https://doi.org/10.1145/3746252.3760790},
doi = {10.1145/3746252.3760790},
booktitle = {Proceedings of the 34th ACM International Conference on Information and Knowledge Management},
pages = {5099–5104},
numpages = {6},
location = {Seoul, Republic of Korea},
series = {CIKM '25}
}

@misc{rajpurkar2017cardiologistlevelarrhythmiadetectionconvolutional,
      title={Cardiologist-Level Arrhythmia Detection with Convolutional Neural Networks}, 
      author={Pranav Rajpurkar and Awni Y. Hannun and Masoumeh Haghpanahi and Codie Bourn and Andrew Y. Ng},
      year={2017},
      eprint={1707.01836},
      archivePrefix={arXiv},
      primaryClass={cs.CV},
      url={https://arxiv.org/abs/1707.01836}, 
}

@misc{oh2026ecgreasoningbenchmarkbenchmarkevaluatingclinical,
      title={ECG-Reasoning-Benchmark: A Benchmark for Evaluating Clinical Reasoning Capabilities in ECG Interpretation}, 
      author={Jungwoo Oh and Hyunseung Chung and Junhee Lee and Min-Gyu Kim and Hangyul Yoon and Ki Seong Lee and Youngchae Lee and Muhan Yeo and Edward Choi},
      year={2026},
      eprint={2603.14326},
      archivePrefix={arXiv},
      primaryClass={cs.LG},
      url={https://arxiv.org/abs/2603.14326}, 
}
